\documentclass[10pt,twocolumn,letterpaper]{article}

\usepackage{cvpr}
\usepackage{times}
\usepackage{epsfig}
\usepackage{graphicx}
\usepackage{subfigure}
\usepackage{amsmath}
\usepackage{amssymb}
\usepackage{booktabs}
\usepackage{diagbox} 
\usepackage{url}

\usepackage[pagebackref=true,breaklinks=true,bookmarks=false, colorlinks]{hyperref}
\usepackage[hyphenbreaks]{breakurl}
\cvprfinalcopy 

\def\cvprPaperID{8662} 

\ifcvprfinal\fi
 \pdfoutput=1
\begin{document}

\title{Learning to Segment the Tail}

\author{
Xinting Hu\textsuperscript{1},\quad Yi Jiang\textsuperscript{2},\quad Kaihua Tang\textsuperscript{1},\quad Jingyuan Chen\textsuperscript{3}, \quad Chunyan Miao\textsuperscript{1},\quad Hanwang Zhang\textsuperscript{1}\\
{\small \textsuperscript{1}Nanyang Technological University,\quad \textsuperscript{2}Alibaba Group,\quad \textsuperscript{3}Damo Academy, Alibaba Group}\\
{\tt \small xinting001@e.ntu.edu.sg, \quad jiangyi0425@gmail.com, \quad kaihua001@e.ntu.edu.sg}\\
{\tt \small jingyuanchen91@gmail.com , \quad ascymiao@ntu.edu.sg, \quad hanwangzhang@ntu.edu.sg}\\}

\maketitle
\thispagestyle{empty}

\begin{abstract}

Real-world visual recognition requires handling the extreme sample imbalance in large-scale long-tailed data. We propose a ``divide\&conquer'' strategy for the challenging LVIS task: divide the whole data into balanced parts and then apply incremental learning to conquer each one. This derives a novel learning paradigm: \textbf{class-incremental few-shot learning}, which is especially effective for the challenge evolving over time: 1) the class imbalance among the old-class knowledge review and 2) the few-shot data in new-class learning. We call our approach \textbf{Learning to Segment the Tail} (LST). In particular, we design an instance-level balanced replay scheme, which is a memory-efficient approximation to balance the instance-level samples from the old-class images. We also propose to use a meta-module for new-class learning, where the module parameters are shared across incremental phases, gaining the learning-to-learn knowledge incrementally, from the data-rich head to the data-poor tail. We empirically show that: at the expense of a little sacrifice of head-class forgetting, we can gain a significant 8.3\% AP improvement for the tail classes with less than 10 instances, achieving an overall 2.0\% AP boost for the whole 1,230 classes\footnote{Code is available at \url{https://github.com/JoyHuYY1412/LST_LVIS}}.  

\end{abstract}
\section{Introduction}
The long-tail distribution inherently exists in our visual world, where a few head classes occupy most of the instances~\cite{intro1,intro2,intro3,intro4}. This is inevitable when we are interested in modeling large-scale datasets, because the class observational probability in nature follows Zipf's law~\cite{zipf}. Therefore, it is prohibitively expensive to counter the nature and collect a balanced sample-rich large-scale dataset, catering for training a robust visual recognition system using the prevailing models~\cite{mrcnn,fastrcnn,fasterrcnn,htc}. 
\begin{figure}[t]
\begin{center}
  \includegraphics[width=1.0\linewidth]{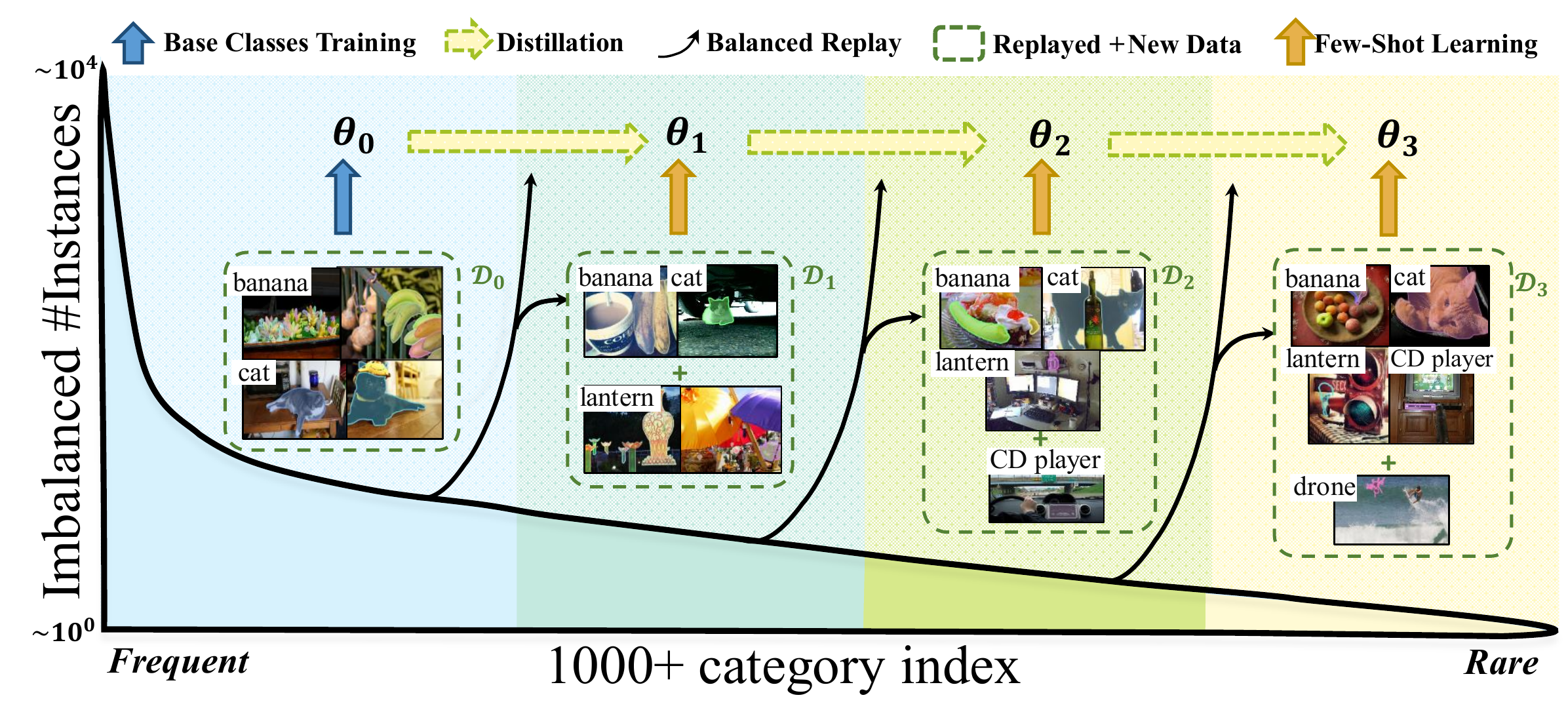}
\end{center}
   \caption{\textbf{Overview of the proposed \textit{Learning to Segment the Tail} (LST) method for LVIS~\cite{lvis}.} To tackle the severe imbalance, we divide the overall dataset into balanced sub-parts $\mathcal{D}_i$ and train the instance segmentation model $\theta_i$ phase-by-phase incrementally. We use knowledge distillation and the proposed \emph{balanced replay} to confront the catastrophic forgetting~\cite{forget} in the fewer- and fewer-shot learning over time. Here, $\theta_3$ is the resultant model.}
 \vspace{-0.2in}
\label{fig:data-1}
\end{figure}

In this paper, we study a practical large-scale visual recognition task on the challenging real-world dataset: Large Vocabulary Instance Segmentation (LVIS)~\cite{lvis}.  As shown in Figure \ref{fig:data-1}, across the 1k+ instance object classes, the number of training instances per class drops from thousands in the head to only a few in the tail (\ie, 26k+ ``banana'' \vs only 1 ``drone''). Empirical studies show that the models trained using such a long-tailed dataset tend to please the common classes but neglect the rare ones~\cite{lvis}. The reasons are two-fold: 1) class imbalance causes the head classes trained thousand times more than the tail classes, and 2) the few-shot samples in the long tail render the generalization a great challenge (\ie, around 300 classes with less than 10 samples). Therefore, the key solution for LVIS is to well address not only the \textbf{imbalance} but also the \textbf{few-shot learning} at a large scale. 

Unfortunately, conventional works on either ``imbalance'' or ``few-shot'' are fundamentally not scalable to LVIS. On the one hand, it is well-known that works on data re-sampling~\cite{resample1,resample2, resample3} --- up-sampling the rare tail classes or down-sampling the frequent head classes --- can prevent the training from being dominated by the head. Nonetheless, as they do not introduce any new diversity, they struggle in the trade-off between the tail over-fit --- the heavy repetitions of the few-shot samples, and the head under-fit --- the significant abandon of the many-shot samples. On the other hand, conventional few-shot learning that transfers the model from a data-rich ``base set'' to a data-poor ``novel set''~\cite{fs5,fs6}, however, is not yet practical in LVIS, as any base or novel split will be eventually imbalanced due to the scale, undermining the generalization ability that is already challenging in few-shot learning~\cite{ls5}. Besides, the scale also raises major memory issues in the episodic training~\cite{matching-net} adopted by recent meta-learning based methods~\cite{fw, fs7}.

An intuitive strategy to address the scale is to \textit{divide} the large ``body'' into ``parts'', \textit{conquer} each of them, and then \textit{merge} them incrementally. As illustrated in Figure~\ref{fig:data-1}, each subset is more balanced and easier to handle. Essentially, the ``divide\&conquer'' strategy for LVIS poses a novel learning paradigm: \textbf{class-incremental few-shot learning}. However, the merge to stitch the parts back to a whole is no longer a trivial adoption of any off-the-shelf class-incremental learning method~\cite{ic5, ic1}. The reason is that different from traditional class-incremental learning scenarios, our incremental phases over time, will face 1) more imbalanced data of the old classes and 2) fewer data of the new classes. This leaves the network more vulnerable to ``catastrophic forgetting''~\cite{forget} in learning the new classes, not to mention the fact that they are fewer- and fewer-shot.

To implement the novel paradigm for the LVIS task, we propose the \emph{balanced replay} scheme for knowledge review and the meta-learning based \emph{weight-generator} module for fast few-shot adaptation. We call our approach: Learning to Segment the Tail (LST). In a nutshell, LST can be summarized in Algorithm~\ref{alg:1}. 
After training the first phase that comes with the abundant labeled data as the bootstrap, we start the incremental learning  in $T$ phases (\eg, $T$ equals 3 in Figure \ref{fig:data-1}). Given the relatively balanced subset $\mathcal{D}_t$ in $t$-th phase using data replay (\texttt{BalancedReplay} in Section~\ref{BalancedReplay}), new classes can be learned and old classes can be fine-tuned simultaneously (\texttt{UpdateModel} in Section~\ref{arch}). To transfer the knowledge step by step from the ``easy'' many-shot head to the ``difficult'' few-shot tail, we furthur adopt a \textit{meta weight generator}~\cite{dlwf} ( \texttt{MWG} in Section~\ref{WeightGen}).

\begin{algorithm}
\caption{Learning to Segment the Tail (\textit{T+1} phases)}
\begin{algorithmic}[1]
    \Require $\left\{\mathcal{G}_i\right\}_{i=0,1,...T}$         \Comment{Dataset pre-processing}
    \newline
    \Ensure  $\theta_T$     \Comment{The final phase model parameters}
    \newline
    \State $\theta_0\!\gets\!\mathop{\arg\min} \limits_{\theta_0}\!L_{inst}(\mathcal{G}_0; \,\theta_0)$ \Comment{Base classes training}
    \For{$t = 1 \to T$}
        \State $\mathcal{D}_t\,\gets\,\Call{{BalancedReplay}}{\left\{\mathcal{G}_i\right\}_{i=0,1,...t}}$;
        \State $\theta_t\,\gets\,\Call{{UpdateModel}}{\mathcal{D}_t,\,\theta_{t-1}}$
    \EndFor
    \newline
    \Function {{UpdateModel}}{$\mathcal{D}_t,\,\theta_{t-1}$}
        \State $\theta_t\,\gets\,\theta_{t-1}$ \Comment{Model initialization} 
        \Repeat
        \If{\textit{use\,meta-module}}  
            \State $\mathcal{G}_t^{sup}\!\gets\! \mathcal{G}_t$ \Comment{Sample support set}
            \State $\theta_t\,\gets\,\Call{{MWG}}{\mathcal{G}_t^{sup}, \theta_t}$
        \EndIf\Comment{Few-shot weight generation}
            \State $\theta_t\!\gets\!\mathop{\arg\min} \limits_{\theta_t}\!\left[L_{inst}(\mathcal{D}_t; \!\theta_t)\!+\!L_{_{kd}}(\mathcal{G}_t,\! \theta_{t\!-\!1}\!;\theta_t)\right]$
            
     \Until{converge} \Comment{Old \& new classes fine-tuning}
    \EndFunction
\end{algorithmic}
\label{alg:1}
\end{algorithm}

We validate the proposed LST on the large-scale long-tailed benchmark LVIS, which contains 1,230 entry-level instance categories. Experimental results show that our LST improves the instance segmentation results over the baseline by 7.0$\sim$8.0\% AP on the tail classes while gaining a 2.2\% overall improvement for the whole classes. The results illuminate us a promising direction for tackling the severe class imbalance in long-tailed data:  class-incremental few-shot learning.   

Our contributions can be summarized as follows:
\begin{itemize}
\setlength{\itemsep}{0pt}
\setlength{\parsep}{0pt}
\setlength{\parskip}{0pt}

\item\noindent We are among the first to study the task of large vocabulary instance segmentation, which is of high practical value by focusing on the severe class imbalance and few-shot learning in the field of instance segmentation.

\item\noindent We develop a novel learning paradigm for LVIS: class-incremental few-shot learning.

\item\noindent The proposed Learning to Segment the Tail (LST) for the above paradigm outperforms baseline methods, especially over the tail classes, where the model can adapt to unseen classes instantly without training.
\end{itemize}

\section{Related Work}
\noindent\textbf{Instance segmentation.} Our instance segmentation backbone is based on the popular region-based frameworks~\cite{inst2,mrcnn,inst3,inst4}, in particular, Mask R-CNN~\cite{mrcnn} and its semi-supervised extension Mask$^X$ R-CNN~\cite{mxrcnn}, which can transfer mask predictor from merely box annotation. However, they cannot scale up for the large-scale long-tailed dataset such as LVIS~\cite{lvis}, which is the focus of our work.

\noindent\textbf{Imbalanced classification.} Re-sampling and re-weighting are the two major efforts to tackle the class imbalance. The former aims to re-balance the training samples across classes~\cite{ic2,resample2,resample3,effective-num}; while the latter focuses on assigning different weights to adjust the loss function~\cite{lt4,reweight1, reweight2,reweight3}. Some works on generalized few-shot learning~\cite{gfsl1,gfsl2} also deal with an extremely imbalanced dataset, extending the test label space of few-shot learning to both base and novel rare classes. We propose a novel re-sampling strategy. Different from previous works that perform on image-level re-sampling, we address the imbalance of dataset on instance-level.

\noindent\textbf{Learning without forgetting \& learning to learn.} Existing works mainly focus on how to learn new knowledge with less forgetting, and how to generalize from the learning process, \ie, learning to learn. To cope with the ever-evolving data, class-incremental learning methods ~\cite{ic1,ic2,ic3,ic4} adapt the original model trained on old classes to new classes, where knowledge distillation~\cite{distill,lwf} and old data replay~\cite{ic5, liu2020mnemonics} are applied to minimize the forgetting. For few-shot learning, meta-learning based works transfer the learning-to-learn knowledge through feature representation~\cite{fs7, meta-weights, fw}, classifier weights~\cite{gfsl1, dlwf}, and the regression of model parameters~\cite{ls5, fs6} from the data-rich base classes,  to obtain a good model initialization for the data-poor new classes. We propose a class-incremental few-shot learning paradigm that can be seen as a non-trivial combination of these two fields.

\section{Learning to Segment the Tail}\label{method}

LVIS is a \textbf{L}arge \textbf{V}ocabulary \textbf{I}nstance \textbf{S}egmentation dataset, which contains 1,230 instance classes~\cite{lvis}. The number of images per class in LVIS has a natural long-tail distribution, with 700+ classes containing less than 100 training samples. To tackle the challenging dataset in the proposed LST using the ``divide\&conquer'' strategy, we first present the division method in Section~\ref{3.1}, and discuss our class-incremental learning pipeline in Section~\ref{arch}. In Section~\ref{BalancedReplay} and Section~\ref{WeightGen}, we detail how to use \texttt{BalancedReplay} and \texttt{MWG} for knowledge review and few-shot adaptation.

\subsection{Dataset Pre-processing} \label{3.1}

Our guideline for the division is to alleviate the intra-phase imbalance of the dataset, where each of division is relatively balanced. We first sort the classes by the number of instance-level samples in a descending order, obtaining a sorted class set $\mathcal{C}$. Then we divide the sorted categories into mutually exclusive groups $\{\mathcal{C}_i\}$. Correspondingly, we have a sub-dataset $\mathcal{G}_i$ with images and annotations for each $\mathcal{C}_i$.

Specifically, after grouping the sorted top $b$ classes as the bootstrap group $\mathcal{C}_0$, and splitting the remaining classes into $T$ evenly spaced bins $\left\{\mathcal{C}_i\right\}_{i=1,2,...T}$,  we obtain the sorted class sets with groups $\mathcal{C}$ = $\mathcal{C}_0 \cup \mathcal{C}_1\cup \cdots \cup \mathcal{C}_{_T}$. By assigning data to the corresponding groups, we convert the whole dataset into $\left\{\mathcal{G}_i\right\}_{i=0,1,...T}$, as shown in line~1 of Algorithm~\ref{alg:1}, where each $\mathcal{G}_i$ is composed of all the annotated images containing any instance of $\mathcal{C}_i$. Following this setting, the data is fed to the network step-wisely, so that our model is trained in a class-incremental learning style.

\begin{figure}[t]
\setlength{\abovecaptionskip}{0.cm}
\setlength{\belowcaptionskip}{-0.cm}
\begin{center}
  \includegraphics[width=1.0\linewidth]{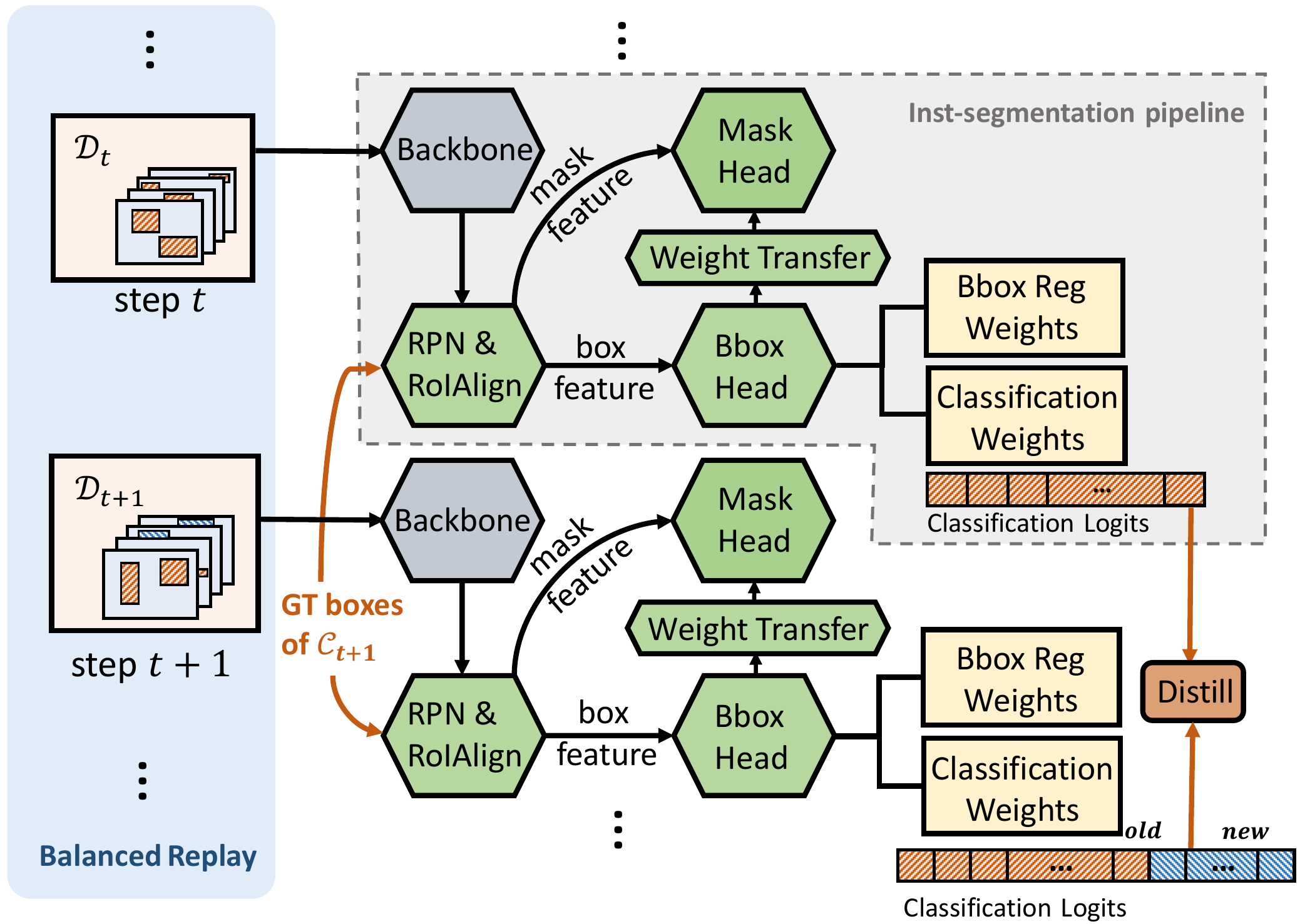}
\end{center}
  \caption{Overview of our framework for learning the instance segmentation model incrementally. It is based on the two-stage instance segmentation architecture, training the overall imbalanced dataset in incremental phases with sampled data for both old and new classes. In incremental phases, the weights of backbone are frozen, and the distillation is calculated using ground truth box annotations between the classification logits of the current and previous networks to avoid forgetting.}
   \vspace{-0.15in}
\label{fig:arch1}
\end{figure}

\subsection{Class-Incremental Instance Segmentation}  
\label{arch}
Class-incremental learning aims to learn a unified model that can recognize  classes of both previous and current phase~\cite{ic5}. In our scenario, we aim to train our network on $\left\{\mathcal{G}_i\right\}_{i=0,1,...T}$, obtaining models from $\theta_0$ to $\theta_T$, and finally deliver $\theta_T$ as our resultant model that can detect all instance classes in LVIS.  Here, we adopt the popular definition inherited from works in incremental learning and few-shot learning~\cite{dlwf,ic1}: classes in $\mathcal{C}_0$ are termed as \textbf{base} classes; for phase $t=1,2,\cdots,T$, classes in $\left\{\mathcal{C}_i\right\}_{i=0,1,...t-1}$ are called \textbf{old} classes and classes in current $\mathcal{C}_t$ are called \textbf{new} classes. For training and evaluation in each phase $t$ , we will not handle anything in the future classes  $\left\{\mathcal{C}_i\right\}_{i=t+1, ..., T}$. As phases $t$ goes by, the data in $\mathcal{G}_t$ for new classes becomes fewer and fewer, and the data for old classes become more and more imbalanced. To tackle the inter-phase imbalance, we propose a novel sampling scheme for the old data, which will be discussed in Section~\ref{BalancedReplay}.

Our overall architecture is shown in Figure~\ref{fig:arch1}. We build our class-incremental learning framework based on Mask$^X$\,R-CNN~\cite{mxrcnn}, which is a modified version of Mask\,R-CNN~\cite{mrcnn}. Mask$^X$\,R-CNN is an instance segmentation model that can be used in partially supervised domain by obtaining a category’s mask parameters from its bounding box parameters. We adopted this weight transfer module so that the class-agnostic transfer function weights can be shared between incremental phases, which can 1) alleviate the training burden for massive mask layers of 1,230 classes and 2) avoid the unstable knowledge distillation of the $28\times 28$ mask logits across classes (\ie, $28\times 28$ times more compared to the class logits distillation in Eq.~\eqref{equ:distill}). Besides, we replaced the last classifier layer in the detection branch with scaled cosine similarity operator, because it has been shown effective in eliminating the bias caused by the significant variance in magnitudes~\cite{ic5,dlwf,ic2}. Formally, given the feature vector $\boldsymbol{x}$, the output logits vector $\boldsymbol{y}$ of cosine similarity classifier with weights $\boldsymbol{w}$ is:
\begin{equation}
\boldsymbol{y} = \overline{\boldsymbol{w}}^T\overline{\boldsymbol{x}}
\end{equation}
where $\overline{\boldsymbol{w}} = \boldsymbol{w} / \left \|\boldsymbol{w}\right \|$ and $\overline{\boldsymbol{x}} = \boldsymbol{x} / \left \|\boldsymbol{x}\right \|$ are the L2-Normalized vectors.

Then the class-specific mask weights in the mask branch are generated from $\boldsymbol{w}$ using the class-agnostic weight prediction function in Mask$^X$\,R-CNN~\cite{mxrcnn}.

The overall class-incremental learning pipeline is shown in Algorithm~\ref{alg:1}, and it is composed of two stages:

\noindent\textbf{Stage 1. Base classes training.}
This training phase ($t=0$) delivers the model $\theta_0$ for base classes, where the backbone and RoI heads are jointly trained. The trained classification weight vectors for top $b$ classes are denoted as $\boldsymbol{W}^{\mathcal{B}}=[\boldsymbol{w}_1,\boldsymbol{w}_2,...\boldsymbol{w}_b]$. We assume that if the data in base classes are sufficiently abundant and relatively balanced, the training of $\theta_0$ can work effectively as the bootstrap for the whole system. 

We calculate the instance segmentation loss $L_{inst}=L_{_{RPN}}+L_{cls} + L_{box} + L_{mask}$. The RPN loss $L_{_{RPN}}$, classification loss $L_{cls}$, bounding-box loss $L_{box}$ and mask loss $L_{mask}$ are identical as those defined in Fast R-CNN~\cite{fastrcnn} and Mask R-CNN~\cite{mrcnn}. 

\begin{figure}[t]
\begin{center}
   \includegraphics[width=1.0\linewidth]{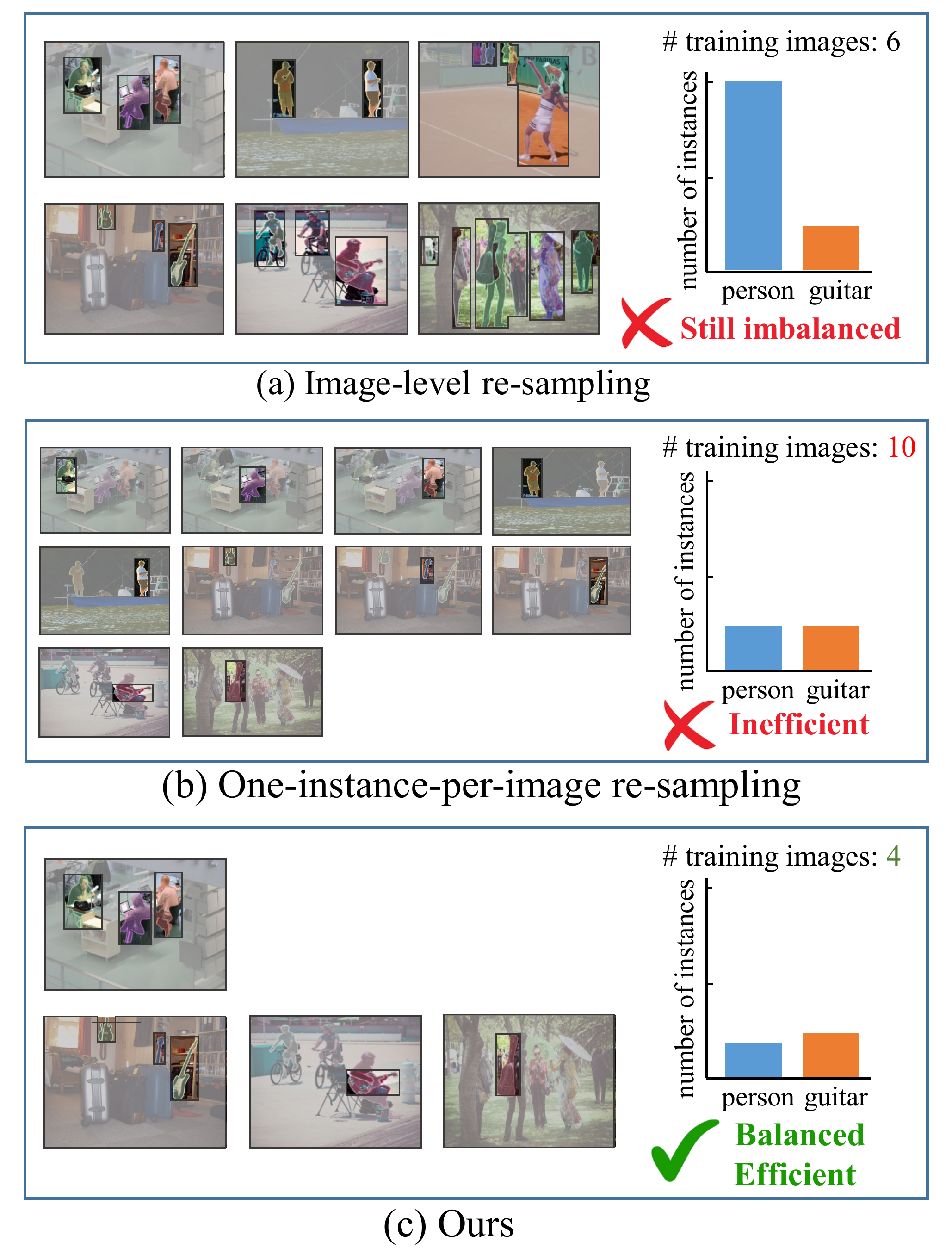}
\end{center}
   \caption{A running example of different re-sampling strategies. Given images of ``person'' and ``guitar'' from different phases, we show the observable instances for each image in training ROI heads using different re-sampling strategies. As shown in (c), compared to (a) and (b), by omitting the annotations of ``person'' in images except for the ones we sampled, our instance-level balanced replay can construct a relatively balanced dataset with much less computation overhead. }
\label{fig:ss_1}
 \vspace{-0.2in}
\end{figure}

\noindent\textbf{Stage 2. Class-incremental learning.} In each phase t (from $1$ to $T$), the number of classifiers is expanded, which leads to the following adjustments to the training procedure in Stage 1:

\noindent\textbf{\textit{Network Expansion.}} After initialized from the last phase's model $\theta_{t-1}$, the current model needs to grow for recruiting new class-specific layers, \ie, the bounding-box, classification and regression layers and the mask prediction layer for new classes. Recall our modifications of the backbone, the weights of mask layers can be transferred from the weights of box layers, so the expansion of the network is only implemented on the box head.

\begin{figure*}
\begin{center}
\includegraphics[width=0.95\linewidth]{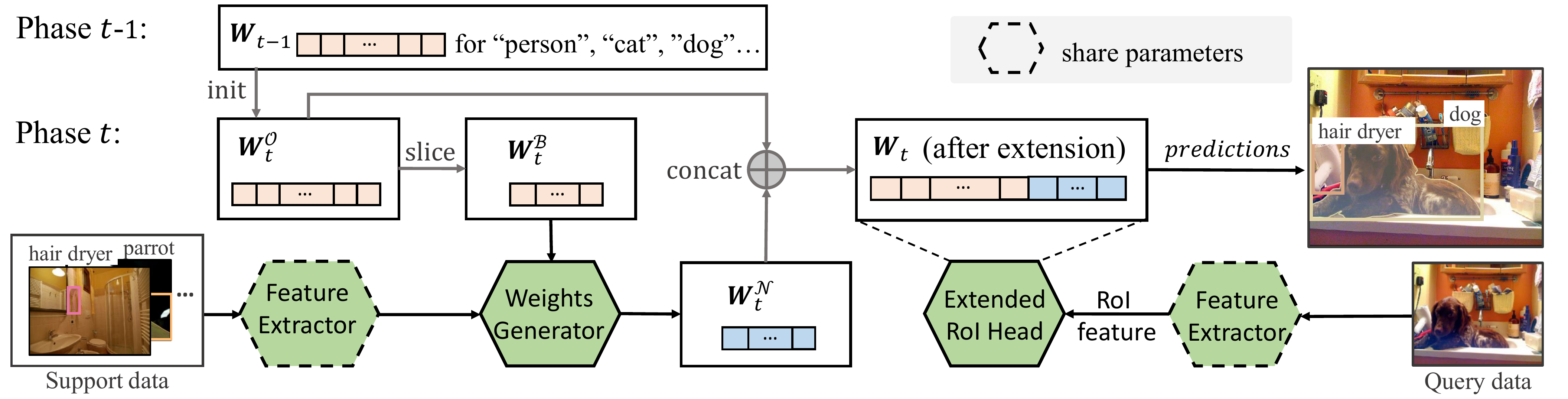}
\end{center}
  \caption{Architecture of our framework combined with weight generator. At the start of each step, classifier weights for old classes are copied from the previous network. Based on the features of new classes samples and base classifier weights, weights for new classes are predicted by our weight generator and concatenated. After obtaining the whole classifier, our weight generator is jointly trained with the network using input image on both old and new categories.}
\label{fig:meta}
\vspace{-0.15in}
\end{figure*}

\noindent\textbf{\textit{Freezing} and \textit{knowledge distillation.}} As discussed in class-incremental learning works~\cite{ic5,ic2}, these two strategies are broadly used to address catastrophic forgetting, the significant performance drop on previous data when adapting a model to new data. Data rehearsal~\cite{ic5} is another strategy to prevent forgetting by reviewing old data, which is discussed in Section~\ref{BalancedReplay}. In our scenario, 1) by \textit{freezing} the weights in the backbone, a strong constraint on the previous representation is imposed, 2) by \textit{knowledge distillation}, the discriminative representation learned previously is not shifted severely during the new learning step. Our distillation loss is defined as:
\begin{equation} \label{equ:distill}
L_{_{kd}}= \left \| \boldsymbol{y}_{t-1} - \boldsymbol{y}_{t}' \right \|
\end{equation}
where $\boldsymbol{y}_{t-1}$ and $\boldsymbol{y}_{t}'$ are the output logits vectors for classes in $\left\{\mathcal{C}_i\right\}_{i=0,1,...t-1}$ using both old model $\theta_{t-1}$ trained in phase $t-1$ and current $\theta_t$, respectively. Note that the output $\boldsymbol{y}_{t}$ in phase t also incorporates new categories in $\mathcal{C}_t$, we use $\boldsymbol{y}_{t}'$ to indicate the sliced logits only corresponding to previous classes $\left\{\mathcal{C}_i\right\}_{i=0,1,...t-1}$. $\left \|\cdot\right\|$ is the L2-distance to measure the difference between logits. We choose L2-distance here in avoid of the grid search of \textit{temperature} as in conventional distillation loss~\cite{lwf}, thanks to the already normalized logits (\ie, logits lie in the same range $\left[-1,1\right]$) using cosine.

The purpose of Eq.~\eqref{equ:distill} is to let the new model mimic the the old model’s behavior (\ie, generate similar output logits), so that the knowledge learned from old network can be preserved. It is worth noting that distillation requires the same input sample going through old and new networks separately. Different from the classification task, in instance segmentation, proposals are dynamically predicted. To this end, we use the ground truth bounding boxes of novel classes as samples in each step for distillation. Overall, for each incremental phase $t$, knowledge distillation loss is added to the final loss as $L = L_{inst} + L_{{kd}}$.

\subsection{Instance-level Data Balanced Replay} 
\label{BalancedReplay}
As shown in Figure \ref{fig:data-1}, within each incremental phase, the variance of instance number is narrowed. However, the inter-phase imbalance (\ie, the gap in the number of samples among phases) exists, leading to a dilemma: if we replay all the previous data, it will definitely break the balance, introducing the imbalance back to our network; if we discard replay, catastrophic forgetting will happen~\cite{ic5}. 

Moreover, previous re-sampling strategies~\cite{lvis,cas} can not be applied gracefully in the instance-level vision tasks. For image-level re-sampling that regularizes the number of images per category, the inherent class co-occurrence may hinder its effectiveness. For example, in Figure~\ref{fig:ss_1}~(a), as ``guitar'' usually co-occur with ``person''$\footnote{ We use ``person'' to replace ``baby'' used to represent a set of synonymous labels: ``child'', ``boy'', ``girl'', ``man'', ``woman'' and ``human'' in LVIS for readability.}$, the adjustment on the number of ``guitar'' instances will always unnecessarily adjust the number of ``person'' instances at the same time. An alternative one-instance-for-one-image strategy in Figure~\ref{fig:ss_1}~(b) can assure the absolute balance, however, the additional computational cost for feed-forwarding the same image multiple times is tremendous. Based on those observations, we proposed the \textit{Instance-level Data Balanced Replay} strategy. For phase $t$, it works as follows:
\begin{itemize}
\vspace{-0.08in}\item [1)] calculate $\bar{n}_{\mathcal{C}}$: the average number of instances \textbf{over all categories} in set $\mathcal{C}_t$;
\vspace{-0.08in}\item [2)] calculate $\{\bar{n}_k\}$: the average number of instances \textbf{over all images} containing annotations from the corresponding old category $ k \in \{\mathcal{C}_i\}_{i=0,1,...t-1}$;
\vspace{-0.08in}\item [3)] construct the replay set $\mathcal{R}_t$: for each category $k$, randomly sample $\lceil \bar{n}_{\mathcal{C}}/{\bar{n}_k} \rceil$ images from images in $\{\mathcal{G}_i\}_{i=0,...t-1}$ containing category $k$, where only those annotations belonging to category $k$ are considered valid in the training.
\end{itemize}
\vspace{-0.08in}
As illustrated in Figure~\ref{fig:ss_1}~(c), by replaying the balanced set $\mathcal{R}_t$ of old data using the above strategy, we dynamically collects a relatively balanced dataset $\mathcal{D}_t = \mathcal{R}_t\cup \mathcal{G}_t$ in each phase $t$.

\subsection{Meta Weight Generator} \label{WeightGen}

So far, the proposed class-incremental pipeline is able to tackle the intra-\&inter-phase imbalance while preserves the performance of classes from the previous phases. However, the challenge of few-shot learning becomes severe as we approach to the tail classes. Therefore, we adopt a Meta Weight Generator (\texttt{MWG}) module~\cite{gfsl1} as shown in Figure~\ref{fig:meta}, which utilizes the base knowledge learned and inherited from the previous phases to dynamically generate the weight matrix of the current phase. The motivation is: given robust feature backbone and classifiers learned for the base classes (\ie, Stage 1 in Section~\ref{arch}), it is possible learning to directly ``write'' new classifiers for the new classes, based on the new sample feature itself and its similarities to the base classifiers~\cite{dlwf}. For an intuitive example, we can customize a ``drone'' classifier by using a ``drone'' sample feature and how the sample looks like the base classes, \eg, 50\% ``airplane'', 30\% ``fan'', and 20\% ``frisbee''.

Formally, in the $t$-th incremental phase, we decompose the classifier weight matrix $\boldsymbol{W}_t$ into two parts: $\boldsymbol{W}_t^{\mathcal{O}}$, $\boldsymbol{W}_t^{\mathcal{N}}$ for the old and the new classes, respectively. Following the Gidaris\&Komodakis's work~\cite{dlwf}, $\boldsymbol{W}_t^{\mathcal{N}}$ is dynamically generated. In particular, we retrieve the base classifier weights $\boldsymbol{W}_t^{\mathcal{B}}$ from $\boldsymbol{W}_t^{\mathcal{O}}$, then learn how to compose $\boldsymbol{W}_t^{\mathcal{N}}$. Take an image containing RoIs of a new category $c$ as an example, for each RoI feature vector $\boldsymbol{x}$, 1) the feature vector $\boldsymbol{x}$ is fed to an attention kernel function to get the coefficients $\boldsymbol{m}$ as: $\boldsymbol{m} = Att(\boldsymbol{K}, \boldsymbol{V}\boldsymbol{x})$, where $\boldsymbol{m}$ are the weight coefficients used to attend $b$ base classifiers weights $\boldsymbol{W}_t^{\mathcal{B}}$, $\boldsymbol{V}$ is a learnable matrix that transforms $\boldsymbol{x}$ to the query vector, and $\boldsymbol{K}$ is a set of learnable keys (one per base category); 2) the classification weight $\boldsymbol{w}_c$ is first generated for each RoI feature $\boldsymbol{x}$ independently and then averaged over all RoIs of category $c$ in this image as the final predicted weight vector of category $c$. For each RoI feature $\boldsymbol{x}$, the corresponding classifier weight is calculated as:
\begin{equation}
\boldsymbol{w} = \mathbf{a}\odot\mathbf{x} + \mathbf{b} \odot(\mathbf{W}_t^{\mathcal{B}}\mathbf{m}),
\label{eqs1fafd}
\end{equation}
where $\odot$ denotes element-wise multiplication, $\mathbf{a}$ and $\mathbf{b}$ are learnable weight vectors.

For the initialization of the $t$-th phase, $\mathbf{W}_t^{\mathcal{O}}$ is copied from the previous phase $t-1$. For the episodic training~\cite{matching-net}, each episode is composed of a support set and a query set sampled from $\mathcal{D}_t$. The support set is for applying \texttt{MWG} to generate $\mathbf{W}_t^{\mathcal{N}}$ (Eq.~\eqref{eqs1fafd}), and the query set is for collecting loss from the predictions using the full model $\theta_t$: the concatenated classifiers $[\mathbf{W}_t^{\mathcal{O}}, \mathbf{W}_t^{\mathcal{N}}$] as well as other network parameters, and then update $\theta_t$. This joint training assures that the classifier weights and the meta-learner are synchronized in the $t$-th phase. After the episodic training, we set the weights for a novel category $c$ by averaging the predicted weights of all the instances of class $c$ in $\mathcal{D}_t$. Then, the meta-module can be completely detached, and we are ready to deliver the model $\theta_t$.

\begin{table*}
\centering
\scalebox{0.85}
{
\begin{tabular}{l|cccccc|ccc|c} 
\hline
Model    & AP$_{(0,1]}$ &  AP$_{(0,5)}$   & AP$_{\left(0,10\right)}$ & AP$_{\left[10,100\right)}$ & AP$_{\left[100,1000\right)}$ & AP$_{\left[1000,-\right)}$ & AP & AP50 & AP75  &AP$^{bb}$ \\
\midrule
Baseline~\cite{mrcnn} & 0.0 &0.0         & 0.0          & 12.8            & 20.9            & \textbf{28.3} & 17.9 &28.9 & 18.8 & 17.9\\
Modified backbone  & 0.0 & 0.0  & 0.0        & 13.9             & 19.9              & 27.6   & 17.8  & 28.2 & 18.8 & 17.7 \\
\midrule
Class-aware Sampling~\cite{cas} & 0.0&  0.0  & 0.0          & 20.0       & 20.2          & 24.5      & 19.5 & 31.6 & 20.5 & 19.3\\
Repeat-factor Sampling~\cite{lvis} & 4.0 & 0.0  & 2.9         & 19.9            & 21.4      & 27.8     & 20.8 & 33.3 & 22.0 & 20.6\\
\midrule
LST w/o MWG (Ours) & 12.0 &9.3  & \textbf{11.7}  & \textbf{27.1}  & 21.3   & 22.3     & 22.8
& 36.4 & 24.1& 22.3\\

LST w MWG (Ours)  & \textbf{13.6} & \textbf{10.7} & 11.2   & 26.8   & \textbf{21.7}               & 23.0     & \textbf{23.0} 
& \textbf{36.7} & \textbf{24.8} & \textbf{22.6}\\
\hline
 \end{tabular}
}
 \vspace{0.1in}
  \caption{Results of our LST and the comparison with other methods on LVIS val set. All experiments are performed based on ResNet-50-FPN Mask\,R-CNN.
  }
 \label{tab:tb1}
 \vspace{-0.15in}
\end{table*}

\section{Experiments}
We conducted experiments on LVIS~\cite{lvis} using the standard metrics for instance segmentation. AP was calculated across IoU threshold from 0.5 to 0.95 over all categories. AP50 (or AP75) means using an IoU threshold 0.5 (or 0.75) to identify whether a prediction is positive. To better display the results from the head to the tail,  AP$_{(0,1]}$, AP$_{(0,5)}$, AP$_{\left(0,10\right)}$, AP$_{\left[10,100\right)}$, AP$_{\left[100,1000\right)}$, AP$_{\left[1000,-\right)}$ were evaluated for the sets of categories which containing only 1, \textless 5, \textless 10, 10 $\sim$ 100, 100 $\sim$ 1,000 and $\geq$ 1,000 training object instances. AP for object detection was reported as AP$^{bb}$.

\subsection{Implementation Details}
We implemented our architectures and other baselines (\eg, Mask$^X$ R-CNN~\cite{mxrcnn}) on the Mask R-CNN~\cite{mrcnn} code base \texttt{maskrcnn\_benchmark}\footnote{\url{https://github.com/facebookresearch/maskrcnn-benchmark}}. For Section~\ref{arch}, we implemented as follows: 1) mask weights were generated by a class-agnostic MLP mask branch together with the weights transferred from the classifiers of the box head following Hu~\etal~\cite{mxrcnn}; 2) cosine normalization was applied to both of the feature vectors and the classifier weights, to obtain the classification logits. Note that the ReLU non-linearity in the final layer was removed to allow the feature vectors to take both positive and negative values. 

We initialized the scaling factor of cosine similarity as 10. All the models were initialized using the released model pre-trained on COCO~\cite{coco}, and trained by using SGD with 1e-4 weight decay and 0.9 momentum. Each minibatch had 8 training images, and the images were resized to that its shorter edge is 800-pixel. No other augmentation was used except for horizontal flipping. Models were evaluated using the 5k \texttt{val} images. Following Gupta~\etal~\cite{lvis}, we increased the number of detections per image up to top 300 (\vs top 100 for COCO) and reduced the minimum score threshold from the default of 0.05 to 0.0.

For Section~\ref{method}, in Stage 1, we chose $b$\,=\,270, where each of the top $b$ classes has 400+ instances. 512 RoIs were selected per image, and the positive-negative ratio is $1\colon3$. For training the top $b$ classes, the learning rate was set to 0.01 and decayed to 0.001 and 0.0001 after 6 epochs and 8 epochs (10 epochs in total). In Stage 2, we split the rest classes into 6 groups. For each incremental phase, we only sampled 100 proposals per image as the number of valid annotations per image shrinks when adopting our balanced replay strategy. Recall the freezing operation in Section~\ref{arch}, we froze the top 3 layers of ResNet~\cite{resnet} in the backbone in each incremental learning phase. The learning rate was from 0.002 and divided by 10 after 6 epochs (10 epochs in total). More experiments on the choice of $b$ and the number of phases are presented in Section~\ref{ablation}.

\begin{figure}[t]
\begin{center}
  \includegraphics[width=1.0\linewidth]{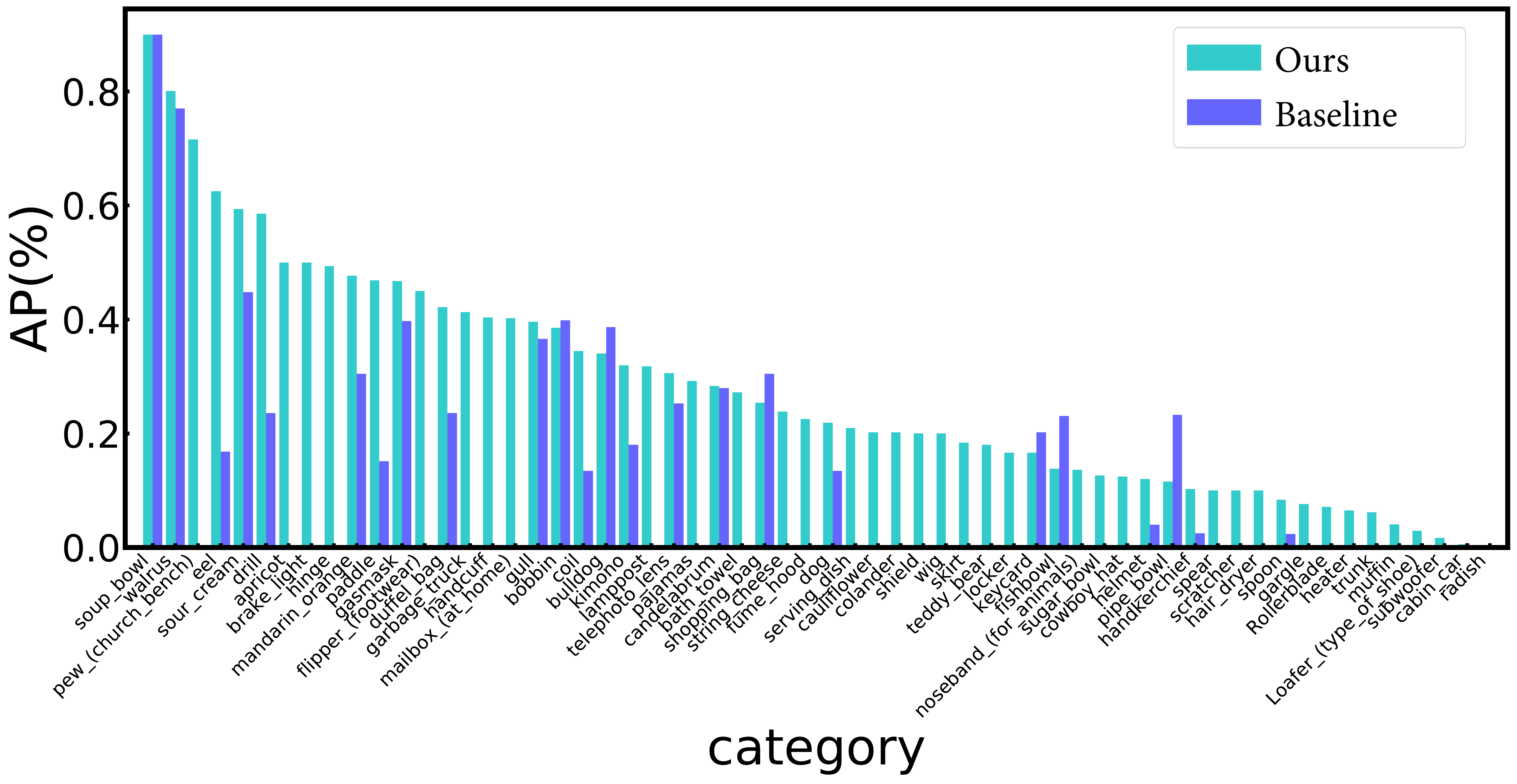}
\end{center}
\vspace{-0.15in}
  \caption{Performance comparison on a subset of tail classes between our LST and the joint training baseline (Mask$^X$ R-CNN). We observe that the baseline APs for many few-shot categories is zero due to the extreme imbalance.}
\label{fig:tail_accu}
\vspace{-0.15in}
\end{figure}

\begin{table}[t]
\centering
\scalebox{0.9}{
\setlength{\tabcolsep}{1mm}
\begin{tabular}{c|c|c|c|c|c}
\hline
Model & AP$_{\left(0,10\right)}$ & AP$_{\left[10,100\right)}$ & AP$_{\left[100,1000\right)}$ & AP$_{\left[1000,-\right)}$ & AP \\
\hline
Baseline & 3.5 & 20.1 & \textbf{25.1} & \textbf{31.5} & 23.0\\
\hline
Ours & \textbf{14.4} & \textbf{30.0} & 25.0 &26.9 & \textbf{26.3}\\
\hline
\end{tabular} 
}
\vspace{0.05in}

\caption{Results of our LST and baseline implemented on ResNeXt-101-32x8d-FPN Mask\,R-CNN.}
\label{resnext}
\vspace{0.01in}
\end{table}

\subsection{Results and Analyses on LVIS}
\begin{figure}[t]
\begin{center}
  \includegraphics[width=0.9\linewidth]{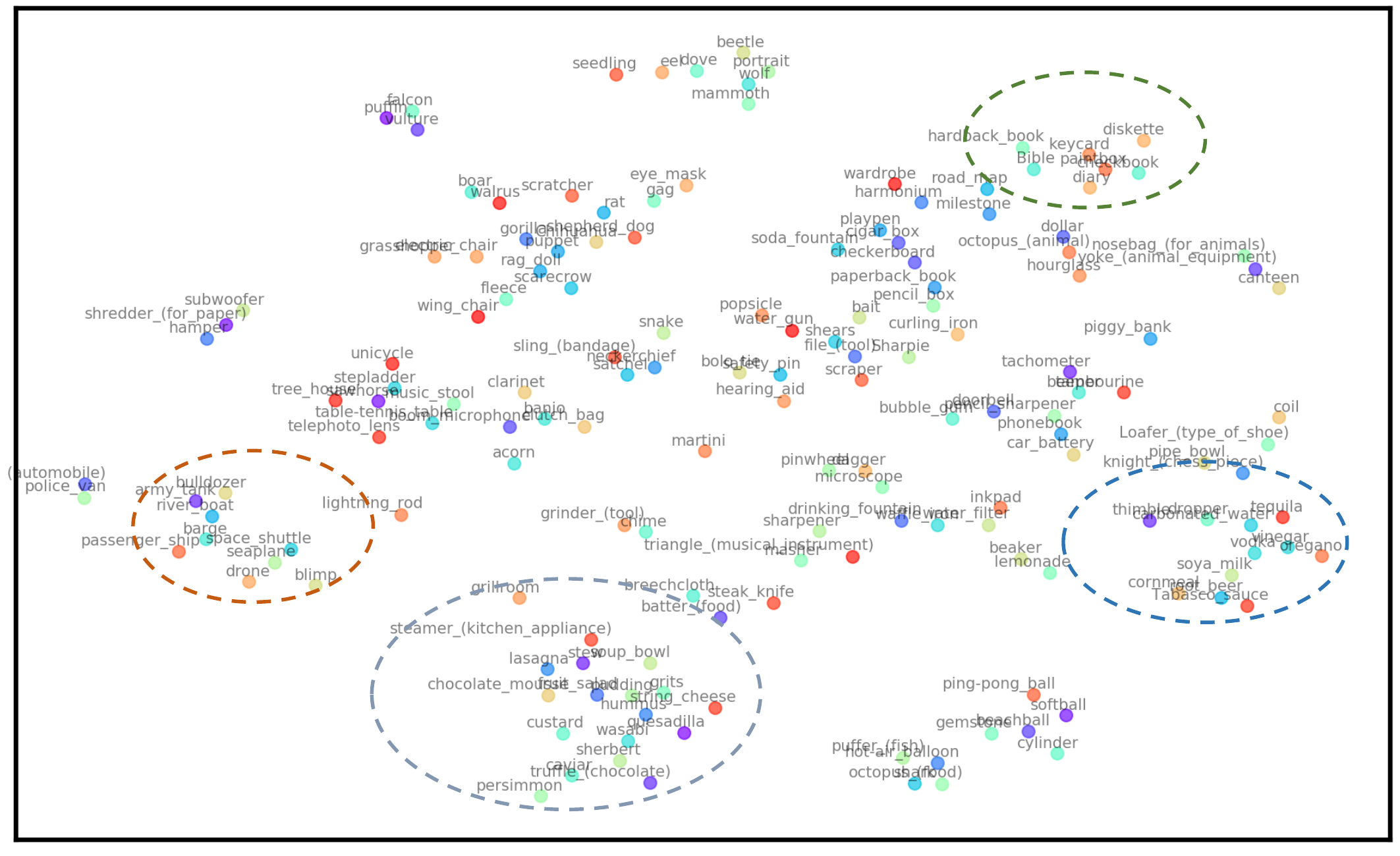}
\end{center}
\vspace{-0.15in}
   \caption{t-SNE~\cite{tsne} embeddings of the coefficients for few-shot categories in the last phase. As noted, semantically and visually similar classes are close (\ie, ``diary'' and ``diskette'', ``custard'' and ``wasabi'').}
\label{fig:tsne}
\vspace{-0.15in}
\end{figure}

\noindent\textbf{Results.}
As shown in Table~\ref{tab:tb1}, our method evaluated at the last phase, \ie, the whole dataset, outperforms the baselines in the tail classes (AP$_{\left(0,10\right)}$ and AP$_{\left[10,100\right)}$) by a large margin. The overall AP for both object detection and instance segmentation improves. Especially, as shown in Figure~\ref{fig:tail_accu}, we randomly sampled 60 classes from the tail classes, whose number of instances in the training set is smaller than 100, and reported the result with and without using our LST which is class-incremental. We observe that our approach obtains remarkable improvement in most tail categories. We also compared our method with other re-sampling methods proposed to tackle the imbalanced data, where \textit{repeat-factor sampling}~\cite{lvis} essentially up-samples the images containing annotations from tail classes, and \textit{class-aware sampling}~\cite{cas} is an alternate oversampling method. The results show that our method surpasses all the other image-level re-sampling approaches on the tail classes, bringing an improvement in overall AP as well. In Figure~\ref{fig:tsne}, we visualized the predicted coefficients vectors $\boldsymbol{m}$ of our weight generator for samples in the last phase. The coefficient vectors of visually or semantically similar classes tend to be close, which shows our weight generator's effectiveness in relating the learning process for data-rich and data-poor classes. Due to limited resources, all the above models were implemented on ResNet-50-FPN. We further report the result applying our method to ResNeXt-101-32x8d-FPN~\cite{resnext} in Table~\ref{resnext} ($b$ = 270, 3 phases), which also shows significant improvement. With more powerful computing resource available, we would like to follow the settings of Tan~\etal's work~\cite{jingru} to further improve our performances. We believe that our findings are regardless of visual backbones and data augmentation tricks.






\noindent
\textbf{Analyses.}
Oksuz~\etal~\cite{imbalance} pointed out that the imbalance among different foreground categories, owing to the dataset itself, undermines the performance of popular recognition models. The results of our baseline models in Table~\ref{tab:tb1} validate this opinion, showing the tendency that the recognition on rare categories performs much worse than the frequent ones (0.0\% \vs 28.3\%) in LVIS. By re-balancing the dataset, previous re-sampling works like Gupta~\etal~\cite{lvis} or Shen~\etal~\cite{cas} somewhat improve the performance for the tail classes. However, we show that they are less effective than our LST. The reason is that they struggle in the trade-off between the tail over-fit and the head under-fit. Furthermore, recall Figure~\ref{fig:ss_1}, our method is more suitable for instance-based tasks as we essentially tackle the overall imbalance over \textbf{instances}. What is more, for Gupta~\etal's work~\cite{lvis}, the threshold used for guiding the re-sampling of the whole dataset is sensitive to the data distribution and thus needs to be carefully tuned. As a result, the method is not flexible when new observations are added to the current dataset, bringing about an expansion of the tail. In contrast, the experiments in Section~\ref{ablation} show that our method is robust to the distribution inside each incremental phase, revealing the potential of our work to be applied to open classes with rarer data.

\subsection{Ablation Study} \label{ablation}
\noindent

\begin{table}[t]
\centering
\scalebox{0.8}{
\setlength{\tabcolsep}{5mm}
\begin{tabular}{c|c|c|c}
\hline
$b$ &  $phase\_size$  & $\# phases$ & AP \\
\hline
110 &  160 & 7 & 22.4\\
\hline
270 & 160 & 6 & 22.8\\
270 & 320 & 3 & 22.9\\
270 & 480 & 2 & 22.9\\
270 & 960 & 1 & 21.8\\
\hline
590 & 160 & 4 & 21.2\\
590 & 320 & 2 & 21.4\\
\hline
\end{tabular} 
}
\vspace{0.1in}

\caption{Ablation study for different size of base classes $b$ and the number of incremental phases.}
\label{tb2}
\vspace{-0.15in}
\end{table}

\begin{figure}[t]
\begin{center}
  \includegraphics[width=1.0\linewidth]{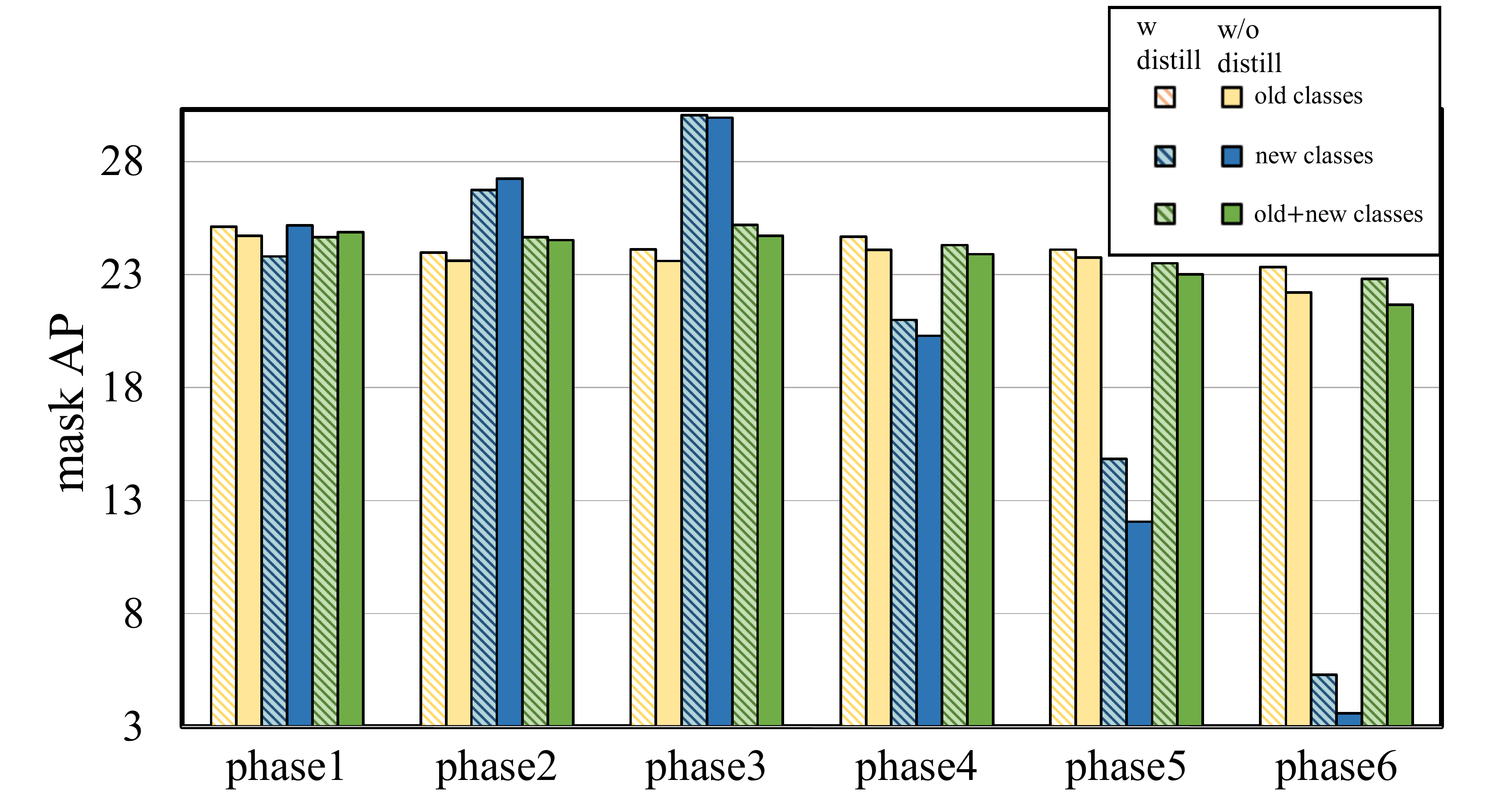}
\end{center}
\vspace{-0.15in}
   \caption{Comparison between networks with using knowledge distillation (shadow fill bars) and without using it (solid fill bars). Results are reported for old classes (\textcolor{yellow}{yellow} bars), new classes (\textcolor{blue}{blue} bars) and old\&new classes (\textcolor{green}{green} bars).}
\label{fig:distill}
\vspace{-0.15in}
\end{figure}

\noindent
\textbf{Choose of $b$ and the size of phase.}
The influence of different $b$ and the number of phases is shown in Table~\ref{tb2}. We empirically show that, on the one hand, the final performance is sensitive to the choice of $b$, as the training on the more imbalanced base dataset (\ie, $b$ = 590) undermines the reliability of $\theta_0$ and further influences the following phases. On the other hand, the results are relatively robust to the size of each incremental phase, as the balanced replay can always provide a relatively balanced dataset when $phase\_size$ locates in a moderate range.


\noindent
\textbf{Knowledge distillation.} We split the rest 960 classes into 6 phases, and examined the influence of using knowledge distillation in each phase by comparing the performances on new classes, old classes, new\&old classes, respectively. As shown in Figure\,\ref{fig:distill}, models trained without distilling classification logits of two adjacent phases perform consistently worse than the model using the distillation on new\&old classes. In the first few phases, the performance of new classes without distillation is higher, because it is trivial that when the new-class data is abundant, ``forgetting'' all the old classes are beneficial to focus on the performances of new classes. But, when the number of instances for each category become fewer and fewer, the distillation becomes more important for both new and old classes. The final instance segmentation AP for the whole dataset with and without knowledge distillation is 22.8\% \vs 21.6\%, demonstrating the effectiveness of the distillation.

\begin{figure}[t]
\centering
\subfigure[LVIS-bboxAP (6 phases)]{ 
\centering 
\includegraphics[width=0.45\linewidth]{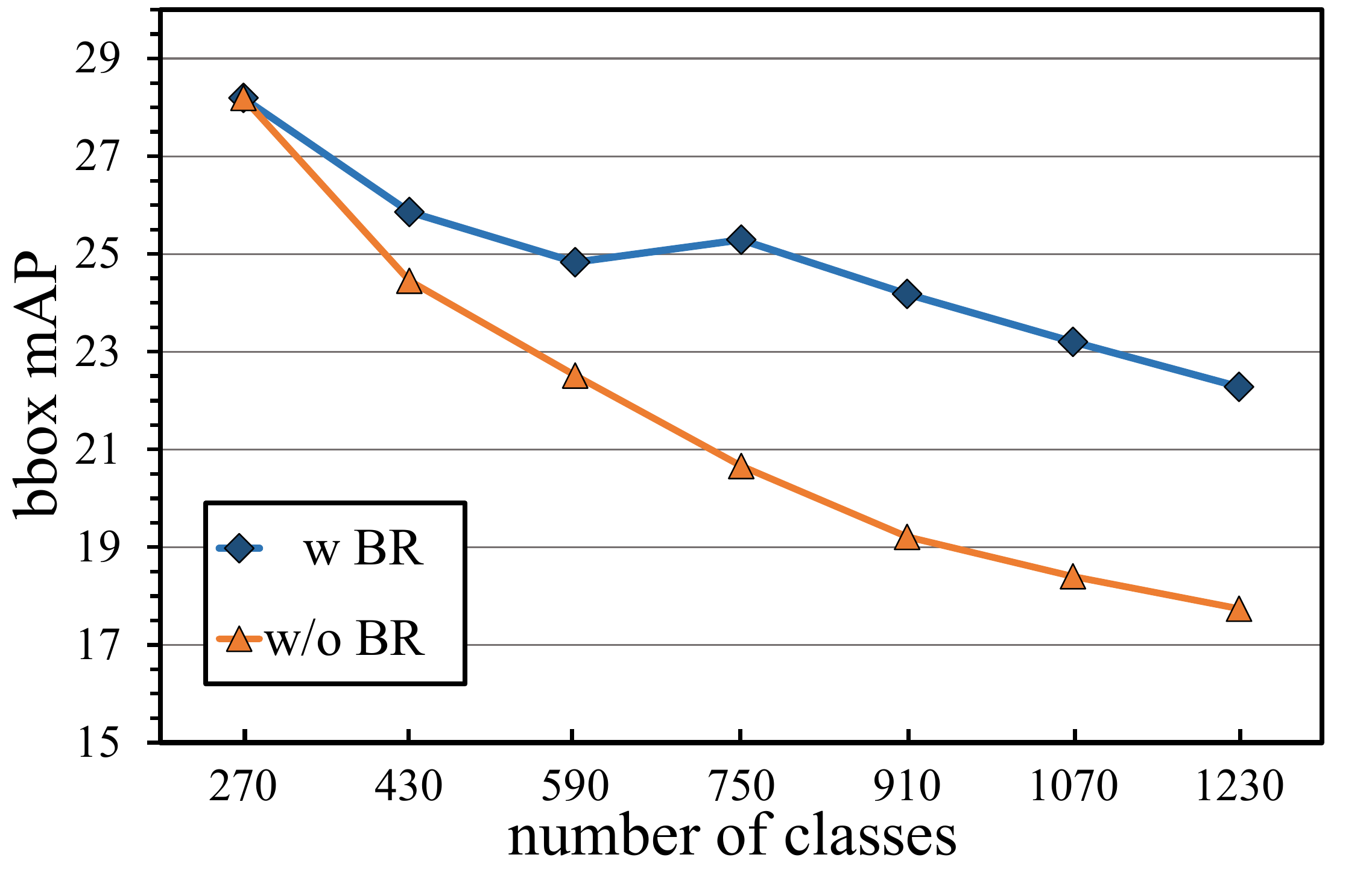} 
}
\subfigure[LVIS-maskAP (6 phases)]{ 
\centering
\includegraphics[width=0.45\linewidth]{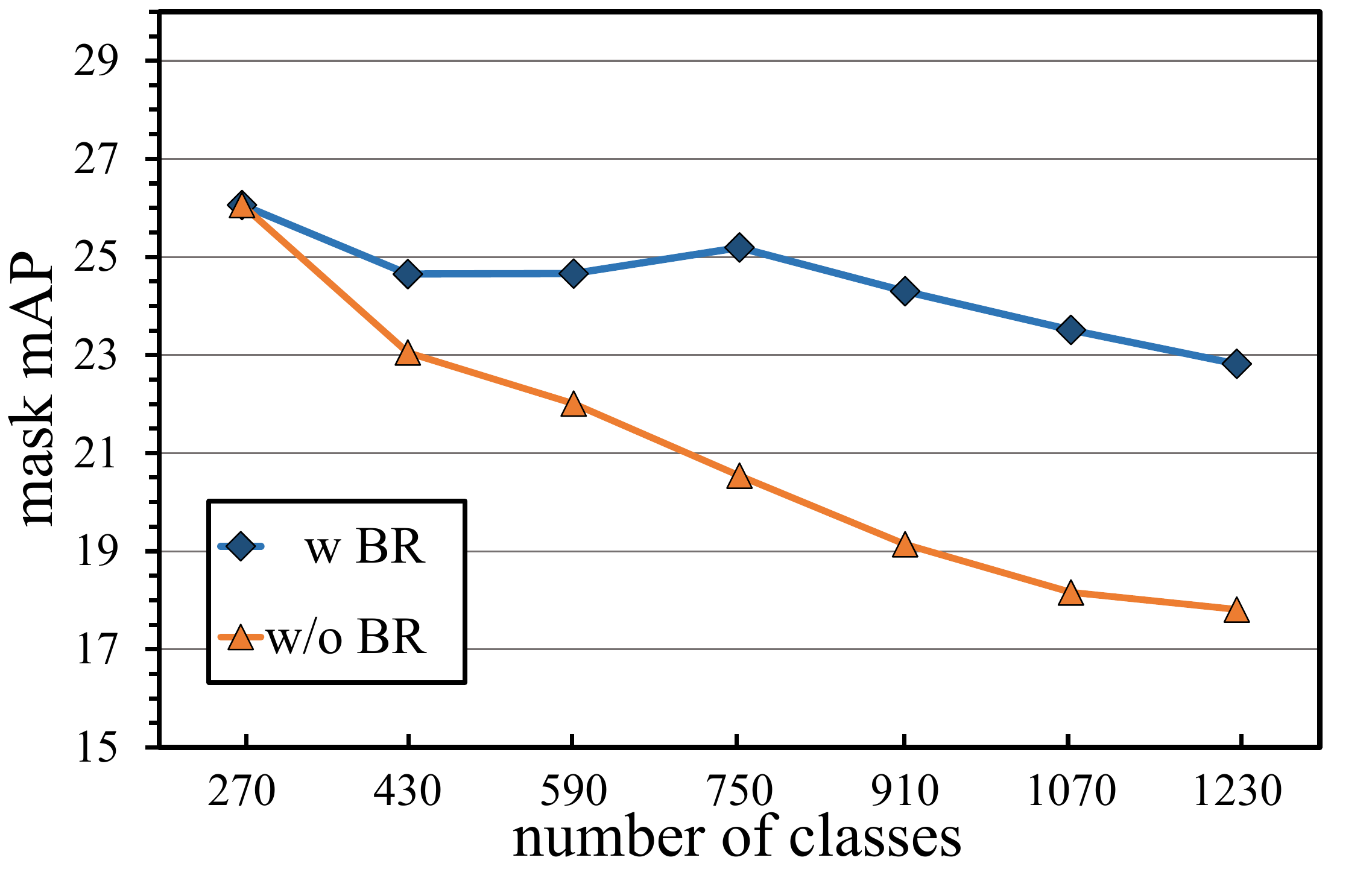} 
}
\caption{Performance comparison for the models trained with and without our balanced replay. For every incremental phase, the detection and instance segmentation performance evaluated on new\&old classes are reported. } 
\label{fig:ss} 
\vspace{-0.05in}
\end{figure}

\noindent
\textbf{Balanced replay.} Figure\,\ref{fig:ss} shows the effect of our Balanced Replay (BR) compared to the baseline that uses all the data from old\&new classes in each phase. It is worth noting that although more data is used for training, the severe imbalance causes the gradually worse performance than our method's. Besides, our method needs far less storage memory consumption and training iterations to converge.

\begin{figure}[t]
\begin{center}
  \includegraphics[width=0.9\linewidth]{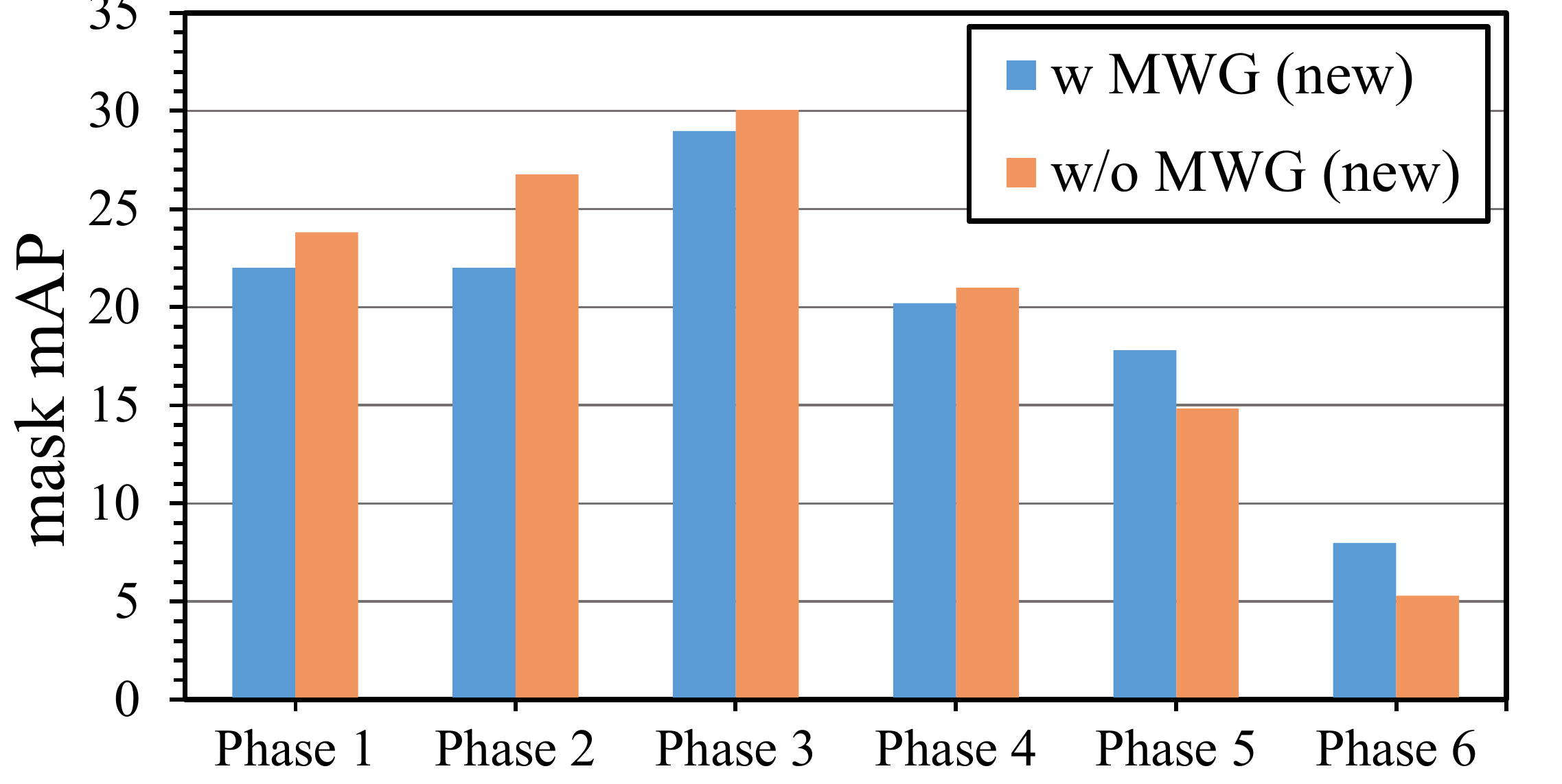}
\end{center}
\vspace{-0.15in}
   \caption{Performance comparison for the models trained with and without our meta weight generator. For every incremental phase, the instance segmentation performance was evaluated on the whole new\&old classes, and we only report the results on the new classes to highlight the few-shot learning performances.}
\label{fig:meta_res}
\vspace{-0.15in}
\end{figure}

\noindent
\textbf{Meta weight generator.} We examined the performance of our system with and without using meta weight generator. As shown in Table~\ref{tab:tb1}, both of them offer a very significant boost on few-shot recognition, while the meta-module based method does better on extreme few-shot classes (\ie, AP$_{(0,1]}$, AP$_{(0,5)}$). More specifically, we evaluated the models at each phase for all classes and report the performance of new classes (Figure~\ref{fig:meta_res}). It is easy to see that among the two, the meta-module based solution exhibits better few-shot recognition behavior, especially for the \textless 5-shot classes in the last phase (5.3\% \vs 8.0\%), without affecting the recognition performance of all classes. However, compared to the conventional training, the episodic training for meta-module is memory-inefficient. In our implementation, 160 is the maximum phase size for network armed with \texttt{MWG}, so we only report the results using 6 incremental phases. We would like to explore a better combination of meta-learning and fine-tuning in future work.
\section{Conclusions}
We addressed the problem of large-scale long-tailed instance segmentation by formulating a novel paradigm: class-incremental few-shot learning, where any large dataset can be divided into groups and incrementally learned from the head to the tail. This paradigm introduces two new challenges over time: 1) for countering the catastrophic forgetting, the old classes are more and more imbalanced, 2) the new classes are more and more few-shot. To this end, we develop the Learning to Segment the Tail (LST) method, equipped with a novel instance-level balanced replay technique and a meta-weight generator for few-shot classes adaptation. Experimental results on the LVIS dataset~\cite{lvis} demonstrated that LST could gain a significant improvement for the tail classes and achieve an overall boost for the whole 1,230 classes. LST offers a novel and practical solution for learning from large-scale long-tailed data: we can use only one downside --- head-class forgetting, to trade off the two challenges --- the large vocabulary and few-shot learning.

\noindent\textbf{Acknowledgements.} We thank all the reviewers for their constructive comments. This work was supported by Alibaba-NTU JRI, and partly supported by Major Scientific Research Project of Zhejiang Lab (No. 2019DB0ZX01).

{\small
\bibliographystyle{ieee_fullname}
\bibliography{egbib}
}

\end{document}